%% file: main.tex
\documentclass[letterpaper, 10 pt, conference]{ieeeconf}
\IEEEoverridecommandlockouts
\input{packages}

\input{notation}

\title{Generative Predictive Control:\\Flow Matching Policies for Dynamic, Difficult-to-Demonstrate Tasks}

\author{Vince Kurtz and Joel W. Burdick}

\begin{document}

\maketitle

\begin{abstract}
    Generative control policies have recently unlocked major progress in robotics. These methods produce action sequences via diffusion or flow matching, with training data provided by demonstrations. But existing methods come with two key limitations: they require expert demonstrations, which can be difficult or costly to obtain, and they are limited to relatively slow, quasi-static tasks. In this paper, we leverage a tight connection between sampling-based predictive control and generative modeling to address these issues. In particular, we introduce \textit{generative predictive control}, a supervised learning framework for tasks with fast dynamics that are easy to simulate but difficult to demonstrate. We show how trained flow-matching policies can be warm-started at inference time, maintaining temporal consistency and enabling high-frequency feedback. We believe that generative predictive control offers a complementary approach to existing behavior cloning methods, and hope that it will pave the way toward generalist policies that extend beyond quasi-static demonstration-oriented tasks. 
\end{abstract}

\section{Introduction and Related Work}\label{sec:intro}

Diffusion and flow matching policies have enabled tremendous success in behavior cloning for quasi-static manipulation \cite{chi2023diffusion, black2024pi_0, zhao2023learning, fu2024mobile}. Can generative policies also control systems with fast nonlinear dynamics at high control frequencies, where demonstrations are difficult to come by? In this paper, we answer this question in the affirmative by introducing generative predictive control (GPC), a supervised learning framework for dynamic and difficult-to-demonstrate tasks. 

Fig.~\ref{fig:hero} summarizes our approach. GPC alternates between data collection via sampling-based predictive control (SPC) and policy training via flow matching. The flow model provides extra samples for SPC, enabling continual performance improvement while maintaining a supervised learning (e.g., regression) objective, which stabilizes the learning process.

\subsubsection{Generative Policies} Diffusion \cite{chi2023diffusion} and flow matching \cite{black2024pi_0} have recently gained prominence as powerful policy representations for robotics. These models typically focus on behavior cloning \cite{zhao2023learning, fu2024mobile}, where expert demonstrations serve as training data. Generative policies have the key advantage of multi-modal expressiveness, allowing multiple ``paths'' to the same goal \cite{chi2023diffusion}. They also provide a natural choice for handling image data \cite{song2019generative,song2020score,lipman2022flow}. Importantly, generative policies are trained in a \textit{supervised} manner, with clearly defined regression targets. This improves training stability over unsupervised reinforcement learning \cite{andrychowicz2020matters, engstrom2019implementation}, but requires demonstrations. Obtaining sufficient demonstration data is a key challenge, particularly for large generalist policies \cite{black2024pi_0, lbm, team2025gemini}. While creative ways to obtain this data are an area of active research \cite{chi2024universal}, it is unlikely that demonstrations alone will produce the internet-scale data used to train large vision-language models any time soon. Furthermore, some tasks are simply difficult to demonstrate, particularly for robots with fast nonlinear dynamics or unique morphologies. 

\subsubsection{Sampling-based Predictive Control (SPC)} In parallel, a very different trend has been gaining traction in the nonlinear optimal control community. SPC is an alternative to gradient-based model predictive control (MPC), where a simple sampling procedure is used in place of a sophisticated nonlinear optimizer \cite{posa2014direct, kurtz2023inverse, aydinoglu2024consensus}. Algorithms in this family include model predictive path integral control (MPPI) \cite{williams2016aggressive}, predictive sampling (PS) \cite{howell2022predictive}, and cross-entropy methods (CEM) \cite{rubinstein1999cross}. SPC algorithms are exceedingly easy to implement, and have been studied for some time. But recent advances in simulation speed and parallelism \cite{mjx, Genesis,makoviychuk2021isaac} allow them to scale to complex problems like dexterous manipulation \cite{li2024drop}, legged locomotion \cite{xue2024full} and more \cite{kurtz2024hydrax, howell2022predictive}. 

\begin{figure}
    \centering
    \includegraphics[width=\linewidth]{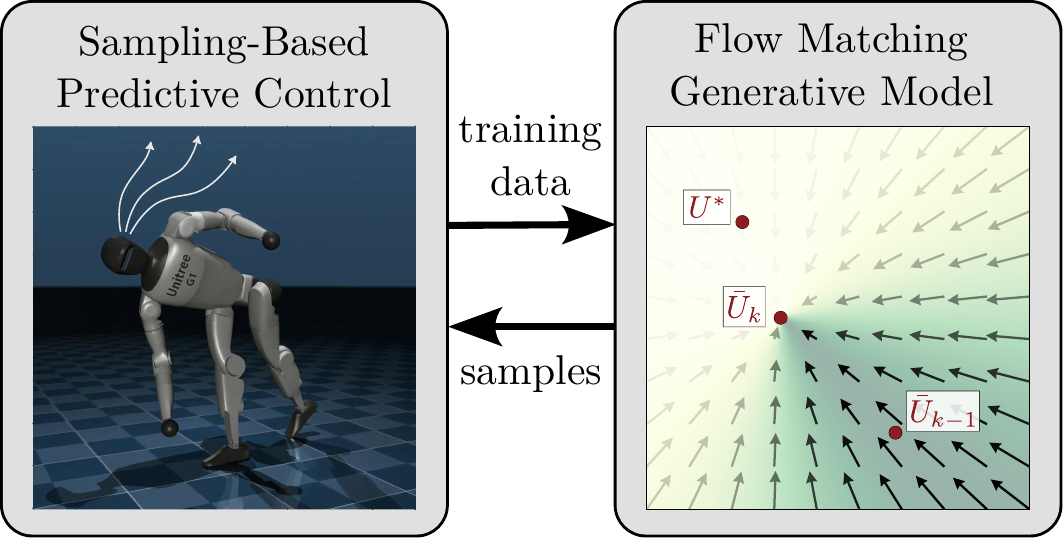}
    \caption{Generative predictive control is a supervised learning framework for dynamic tasks that are difficult to demonstrate but easy to simulate. First, we generate training data with sampling-based predictive control \cite{li2024drop, williams2016aggressive, howell2022predictive}, leveraging advances in massively parallel GPU simulation \cite{mjx, Genesis, makoviychuk2021isaac}. We then use this data to train a flow matching policy, which in turn provides additional high-quality samples. This results in better training data for subsequent iterations, in a virtuous cycle. Code available: \url{https://github.com/vincekurtz/gpc}.}
    \label{fig:hero}
\end{figure}

In many ways, SPC complements generative behavior cloning: it is agnostic to a robot's morphology, and can be quite effective on tasks with fast nonlinear dynamics. Behavior cloning, on the other hand, has shown success on tasks involving deformable objects like cloth and food \cite{chi2023diffusion, black2024pi_0, zhao2023learning, fu2024mobile}, which are hard to simulate at speeds sufficient for SPC. 

\subsubsection{Our Contribution} This paper highlights a deep connection between generative policies and SPC. This connection was first identified by \cite{pan2024model} in the context of MPPI. DIAL-MPC \cite{xue2024full} used this connection to improve SPC performance for legged locomotion, but DIAL-MPC is fully online and does not include policy training. In this paper, we extend this connection to a general class of SPC algorithms and leverage it train flow matching policies. 

In particular, we propose GPC, a supervised learning framework for difficult-to-demonstrate tasks. We show how flow-matching policies can be warm-started to encourage temporal consistency, and demonstrate that warm-starts are critical for high-rate feedback. We show that GPC outperforms proximal policy optimization (PPO) \cite{schulman2017proximal} in many cases, and that our proposed warm-start scheme is more effective than action inpainting \cite{black2025real} at high-frequency control rates. 

To test scalability limits, we train policies on systems ranging from an inverted pendulum to a humanoid robot, and evaluate effectiveness in simulation. Our most difficult task (humanoid standup) exposes current scalability limits: seeding SPC with a GPC policy is effective, but applying the policy directly is not. With this in mind, we provide a detailed discussion of our method's limitations and directions for future work. 

\section{Background}\label{sec:background}

We consider optimal control problems of the standard form
\begin{subequations}\label{eq:ocp}
\begin{align}
    \min_{u_0, u_1, \dots, u_T} ~& \phi(x_{T+1}) + \sum_{\tau=0}^{T} \ell(x_\tau, u_\tau), \\
    \mathrm{s.t.} ~& x_{\tau+1} = f(x_\tau, u_\tau), \\
                   & x_0 = x_{init},
\end{align}
\end{subequations}
where $x_\tau \in \mathbb{R}^n$ represents the system state at time step $\tau$, $u_\tau \in \mathbb{R}^m$ are control actions, $\ell(\cdot, \cdot)$ and $\phi(\cdot)$ are running and terminal costs, and $f(\cdot, \cdot)$ captures the system dynamics. For simplicity, we denote the $T$-length action sequence as $U = [u_0, u_1, \dots, u_T]$ and rewrite \eqref{eq:ocp} in compact form as
\begin{equation}\label{eq:ocp_compact}
    \min_{U} ~ J(U;x_{init}),
\end{equation}
where both the costs and dynamics constraints are incorporated into the (possibly non-convex) objective $J$.

\subsection{Sampling-based Predictive Control}\label{sec:background:spc}

MPC methods traditionally solve $\eqref{eq:ocp}$ with gradient-based non-convex optimization. But these techniques face significant challenges, particularly when it comes to the stiff and highly nonlinear contact dynamics essential for contact-rich robot locomotion and manipulation \cite{wensing2023optimization, posa2014direct, kurtz2023inverse, aydinoglu2024consensus}. 

In response to these challenges, SPC is gaining prominence as a simple and computationally efficient alternative to gradient-based MPC \cite{li2024drop, williams2016aggressive, xue2024full, williams2017model, vlahov2024mppi}. Instead of relying on complex nonlinear optimization, SPC algorithms perform a variation on the following simple procedure:
\begin{enumerate}
    \item At step $k$, sample $N$ candidate action sequences from a Gaussian proposal distribution\footnote{Some SPC methods use a non-isotropic proposal distribution and update the variance of the proposal distribution along with the mean. We focus on an isotropic Gaussian with fixed variance for simplicity.}.
    \begin{equation}\label{eq:spc_proposal}
        U^{(i)} \sim \mathcal{N}(\bar{U}_{k-1}, \sigma^2), \quad i \in [1, N].
    \end{equation}
    \item Roll out (simulate) each action sequence from the latest state estimate $x_{k-1}$, recording the costs
    \begin{equation}\label{eq:spc_rollouts}
        J^{(i)} = J\left(U^{(i)}; x_{k-1}\right).
    \end{equation}
    \item Update the mean action sequence according to some weighting function $g : \mathbb{R} \to \mathbb{R}^+$,
    \begin{equation}\label{eq:spc_update}
        \bar{U}_{k} = \bar{U}_{k-1} + \frac{\sum_{i=1}^{N} g(J^{(i)})(U^{(i)} - \bar{U}_{k-1})}{\sum_{i=1}^N g(J^{(i)})}.
    \end{equation}
    \item Apply the first action from $\bar{U}_{k}$, and repeat in MPC fashion from the updated state $x_k$.
\end{enumerate}

Different SPC algorithms arise from various choices of $g(\cdot)$. For instance, \textbf{MPPI} \cite{williams2016aggressive} uses a Boltzmann-like exponentially weighted average,
\begin{equation}\label{eq:mppi_g}
    g_{MPPI}(J) = \exp\left(- J / \lambda \right),
\end{equation}
where $\lambda > 0$ is the temperature parameter. A smaller $\lambda$ gives more weight to the lowest-cost samples. In the low-temperature limit we recover \textbf{predictive sampling} \cite{howell2022predictive},
\begin{equation}\label{eq:ps_g}
    g_{PS}(J) = \lim_{\lambda \to 0} \exp\left(- J/\lambda\right),
\end{equation}
where the updated mean $\bar{U}$ is simply chosen as the lowest-cost sample. Another popular option is \textbf{CEM} \cite{rubinstein1999cross, li2024drop}, which weighs the top performing samples equally,
\begin{equation}\label{eq:cem_g}
    g_{CEM}(J) = \begin{cases}
        1 & \mathrm{if~} J \leq \gamma \\
        0 & \mathrm{otherwise}
    \end{cases}.
\end{equation}
The threshold $\gamma$ is defined implicitly by a user-selected number of \textit{elite samples}. CEM is typically paired with an adaptive update rule for the variance of the proposal distribution. \textbf{Tsallis-MPPI} \cite{wang2021variational} provides a middle ground between MPPI and CEM via a generalized exponential,
\begin{equation}\label{eq:tsallis_g}
    g_{TMPPI}(J) = \max(1 - (r - 1)J/\lambda, 0)^{\frac{1}{r - 1}},
\end{equation}
where the original MPPI update is recovered as $r \to 1$.



The ability to parallelize rollouts \eqref{eq:spc_rollouts} is a key advantage over gradient-based MPC, an advantage that is further reinforced by massively parallel GPU simulators \cite{mjx, Genesis, makoviychuk2021isaac}. Sampling can also reduce the severity of local minima and smooth out stiff dynamics associated with contact \cite{suh2022bundled, le2024leveraging}.

\subsection{Generative Modeling}\label{sec:background:generative}

Generative modeling considers a seemingly different problem: produce a sample $\x$ from a probability distribution $p(\x)$. In the typical setting, we do not have access to $p(\x)$ in closed form, but we do have samples (training data) from $p(\x)$.

\subsubsection{Flow Matching} Flow matching \cite{lipman2022flow} is based on the idea of a \textit{probability density path} $p_t(\x)$. This path flows from an easy-to-sample distribution $p_0(\x) = \mathcal{N}(0, I)$ at $t = 0$ to the data distribution at $t = 1$. Flow matching methods learn a vector field $v_\theta(\x, t)$ that moves samples along this path.

The flow network $v_\theta$ is trained via standard (stochastic) gradient descent methods on
\begin{equation}\label{eq:flow_objective}
    \min_{\theta} \mathbb{E}_{t, \x_0, \x_1} \left[\mathcal{L}_{FM}(\theta; \x_0, \x_1, t)\right],
\end{equation}
where $\theta$ are learnable parameters (e.g., network weights) and
\begin{equation}\label{eq:flow_loss}
    \mathcal{L}_{FM}\left(\theta; \x_0, \x_1, t \right) = \left\| v_\theta\big(t \x_1 + (1-t)\x_0, t\big) - (\x_1 - \x_0) \right\|^2.
\end{equation}
The expectation in \eqref{eq:flow_objective} is taken over $t \sim \mathcal{U}(0, 1)$, $\x_0 \sim p_0(\x)$, and $\x_1 \sim p_1(\x)$. Each of these are easy to sample, since we already have data points $\x_1$, and sampling $p_0(\x)$ is trivial. Intuitively, $\mathcal{L}_{FM}$ pushes samples in a straight line from $\x_0$ to $\x_1$. At inference time, we first sample $\x \sim p_0(\x)$, then integrate
$\dot{\x} = v_\theta(\x, t)$ from $t = 0$ to $t = 1$, typically with a simple explicit Euler scheme. 

\subsubsection{Diffusion} Flow matching is equivalent (under some technical conditions \cite{gao2025diffusionmeetsflow}) to diffusion-based generative modeling \cite{song2019generative, song2020score}. Diffusion models also consider a series of probability distributions flowing from an initial Gaussian to the data distribution. But rather than being parameterized by a time $t$, these are typically parameterized by noise $\sigma$,
\begin{equation}
    p_\sigma(\x) = \int p(\y) \mathcal{N}(\x;\y, \sigma^2I) d\y.
\end{equation}
For large $\sigma$, $p_\sigma(\x)$ approaches an easy-to-sample Gaussian. For small $\sigma$, $p_\sigma(\x)$ approaches $p(\x)$.

Diffusion models learn the score $s_\theta(\x, \sigma) \approx \nabla_{\x} \log p_\sigma(\x)$ by removing noise added to the original data \cite{song2019generative, song2020score}. We can then use $s_\theta$ to sample from $p_\sigma$ using Langevin dynamics
\begin{equation}
    \x \gets \x + \epsilon s_\theta(\x, \sigma) + \sqrt{2\epsilon} \mathbf{z} \quad \mathbf{z} \sim\mathcal{N}(0, I),
\end{equation}
with step size $\epsilon > 0$. By gradually reducing $\sigma$, we arrive at samples from the data distribution $p(\x)$.

\section{SPC is Online Generative Modeling}

This section establishes a formal connection between SPC and generative modeling: we show that the SPC update \eqref{eq:spc_update} is a Monte Carlo estimate of the score of a noised target distribution. This connection was first identified for the case of MPPI in \cite{pan2024model} and used to develop DIAL-MPC, a multi-stage SPC algorithm for legged robots \cite{xue2024full}. Here we extend this connection to generic SPC algorithms of the form \eqref{eq:spc_update}.

First, we define a state-conditioned target distribution:
\begin{equation}
    p(U \mid x) \propto g(J(U; x)),
\end{equation}
which is determined by the algorithm-specific weighting function $g(\cdot)$ introduced in \eqref{eq:spc_update}. In the spirit of score-based diffusion \cite{song2020score}, we define the noised target distribution
\begin{equation}\label{eq:noised_target}
    p_\sigma(U \mid x) \propto \mathbb{E}_{\tilde{U} \sim \mathcal{N}(U, \sigma^2)}[g(\tilde{U})].
\end{equation}
The score of this noised target is directly used in SPC:

\begin{proposition}
    The score of the noised target distribution \eqref{eq:noised_target} is given by
    \begin{equation}
        \nabla_U \log p_\sigma(U \mid x) = \frac{1}{\sigma^2} \frac{\mathbb{E}_{\tilde{U} \sim \mathcal{N}(U, \sigma^2)}\left[g(\tilde{U})(\tilde{U}-U)\right]}{\mathbb{E}_{\tilde{U} \sim \mathcal{N}(U, \sigma^2)}\left[g(\tilde{U})\right]}.
    \end{equation}
\end{proposition}

\begin{proof}
    For simplicity of notation, we drop the conditioning on $x$ and write the target distribution as $p_\sigma(U)$. We also denote the normal density as
    \[ q_U(\tilde{U}) \triangleq \mathcal{N}(\tilde{U}; U, \sigma^2). \]
    The score of the target distribution is given by
    \begin{equation}
        \nabla_U \log p_\sigma(U) = \frac{\nabla_U p_\sigma(U)}{p_\sigma(U)}.
    \end{equation}
    In the numerator we have
    \begin{align}
        \nabla_U p_\sigma(U) &= \frac{1}{\eta} \nabla_U \int q_U(\tilde{U})g(\tilde{U})d\tilde{U} \\
        &= \frac{1}{\eta} \int \nabla_U q_U(\tilde{U}) g(\tilde{U})d\tilde{U} \\
        &= \frac{1}{\eta} \int q_U(\tilde{U}) \nabla_U \log q_U(\tilde{U})g(\tilde{U})d\tilde{U} \\
        &= \frac{1}{\eta} \mathbb{E}_{\tilde{U} \sim \mathcal{N}(U, \sigma^2)}\left[{g(\tilde{U})\frac{\tilde{U} - U}{\sigma^2}}\right],
    \end{align}
    where $\eta$ is a normalizing constant and we use the fact that $\nabla_U \log q_U(\tilde{U}) = (\tilde{U} - U)/\sigma^2$.

    Bringing $1 / \sigma^2$ outside the expectation, we have
    \begin{equation}
        \frac{\nabla_U p_\sigma(U)}{p_\sigma(U)} = \frac{\mathbb{E}_{\tilde{U} \sim \mathcal{N}(U, \sigma^2)}\left[g(\tilde{U})(\tilde{U}-U)\right]}{\sigma^2\mathbb{E}_{\tilde{U} \sim \mathcal{N}(U, \sigma^2)}\left[g(\tilde{U})\right]}
    \end{equation}
    and thus the proposition holds.
\end{proof}

This means that the SPC update \eqref{eq:spc_update} provides a Monte Carlo estimate of score ascent, e.g.,
\begin{equation}
    \bar{U}_k \gets \bar{U}_{k-1} + \sigma^2 \nabla_{\bar{U}_{k-1}} \log p_\sigma(\bar{U}_{k-1} \mid x_{k-1})\ .
\end{equation}
The additional $\sigma^2$ term may seem like an annoyance, but in fact Langevin step sizes $\epsilon \propto \sigma^2$ are a standard recommendation in the diffusion literature \cite[Algorithm~1]{song2019generative}. Here, the step size choice emerges naturally from the SPC update \eqref{eq:spc_update}.

This connection also sheds light on the benefits of predictive sampling (where we merely choose the best sample) \cite{howell2022predictive}. In particular, the unnoised target distribution for predictive sampling is a Dirac delta concentrating all probability mass at global optima \cite[Appendix~A]{pan2024model}, leading to a noised target distribution $p_\sigma(U \mid x)$ with modes at globally optimal solutions. For this reason, we focus our numerical investigations primarily on predictive sampling. A more thorough exploration of the advantages and disadvantages of other SPC algorithms is an important topic for future work. 

\section{Generative Predictive Control}

The previous section shows that we can think of the mean of the SPC sampling distribution, $\tilde{U}_k$, as being drawn from the state-conditioned optimal action distribution
\begin{equation}\label{eq:gpc_target}
    \bar{U}_k \sim p(U \mid x_k) \propto g\left(J(U; x_k)\right).
\end{equation}
This leads to a natural question: can we train a generative model to produce $\bar{U}_k$ directly? In addition to imitating the SPC update process, such a generative model
\begin{equation}\label{eq:gpc_param_model}
    p_\theta(U \mid x_k) \approx p(U \mid x_k),
\end{equation}
parameterized by network weights $\theta$, would maintain a structure compatible with the flow matching and diffusion models used in behavior cloning \cite{chi2023diffusion, fu2024mobile, zhao2023learning, black2024pi_0}.

\begin{remark}\label{rk:observation_conditioning}
    Behavior cloning methods condition on observations $y = h(x)$ (or a history of observations) rather than a full state estimate \cite{chi2023diffusion}. While we write $p_\theta(U \mid x)$ for notational simplicity, the GPC framework works with observation conditioning. In fact, our implementation uses observations $h(x)$ rather than the full state $x$.
\end{remark}

This is the basic idea behind GPC. We use data ($\bar{U}_k, x_k$) from running SPC in simulation to train a flow matching model \eqref{eq:gpc_param_model}. This model is characterized by a vector field
\begin{equation}
    \dot{U} = v_\theta(U, x, t)
\end{equation}
that pushes samples from $U_t \sim \mathcal{N}(0, I)$ at $t = 0$ to the target distribution \eqref{eq:gpc_target} at $t = 1$. To learn this vector field, we minimize a conditional flow matching loss similar to \eqref{eq:flow_objective},
\begin{multline}\label{eq:gpc_loss}
    \mathcal{L}_{GPC}(\theta; U_0, \bar{U}_k, x_k, t) = \\
     \left\| v_\theta(t\bar{U}_k + (1 - t)U_0, x_k, t) - (\bar{U}_k - U_0) \right\|^2,
\end{multline}
where $(\bar{U}_k, x_k)$ are data points generated by the SPC controller, $U_0 \sim \mathcal{N}(0, I)$ is a sample from the proposal distribution, and $t \sim \mathcal{U}(0, 1)$ is sampled along the path\footnote{To further improve training efficiency, we weigh data points according to cosine similarity with $\bar{U}_k - \bar{U}_{k-1}$. This puts greater emphasis on samples similar to $\bar{U}_{k-1}$, as illustrated by the shading in Fig.~\ref{fig:hero}.}.

However, \textbf{directly training a generative model on SPC data is not particularly effective}, as the training targets are very noisy \cite{zhu2024should}. To avoid this issue, GPC performs several cycles of SPC simulation and model fitting, as illustrated in Fig.~\ref{fig:hero} and outlined in Algorithm~\ref{alg:gpc}. In each cycle, samples from the partially-trained flow matching policy bootstrap SPC, providing an improved sampling distribution and thus better training data for the next model fitting step.

\begin{algorithm}
    \caption{Generative Predictive Control}
    \label{alg:gpc}
    \DontPrintSemicolon
    \KwIn{SPC algorithm $g(U)$, flow matching model $p_\theta(U \mid x)$, system model $f(x, u)$.}
    \KwOut{Trained flow model parameters $\theta$.}
    
    \While{not converged}{
        \For{$j = 1,...,N_E$} { \label{ae:envs}
            Sample initial conditions (parallel envs):
        
            $x^{(j)}_0 \sim \mathcal{X}_0$

            $\bar{U}^{(j)}_0 \sim \mathcal{N}(0, \sigma^2I)$
            
            \For{$k \in [1, K]$}
            {
                Sample action sequences:
                
                $U^{(i,j)} \sim \mathcal{N}(\bar{U}^{(j)}_{k-1}, \sigma^2 I), \quad i \in [1, N_S]$ \label{ae:spc_samples}
                
                $U^{(i,j)} \sim p_\theta(\bar{U}_{k-1}^{(j)} \mid x_{k-1}^{(j)}), \quad i \in [N_S, N]$ \label{ae:flow_samples}
                
                Parallel rollouts:
                
                $J^{(i,j)} \gets J\left(U^{(i,j)}; x^{i,j}_{k-1}\right)$ \label{ae:rollouts}

                Update actions via SPC:

                $\bar{U}_k^{(j)} \gets \bar{U}_{k-1}^{(j)} + \frac{\sum_{i=1}^{N} g(J^{(i,j)})(U^{(i,j)} - \bar{U}^{(j)}_{k-1})}{{\sum_{i=1}^N g(J^{(i,j)})}}$
                
                Advance (parallel) simulations:
                
                $x_{k}^{(j)} \gets f\left(x_{k-1}^{(j)}, u_{k-1}^{(j)}\right)$ \label{ae:sim}

            }
        }

        Fit flow matching model:
        
        $\min_{\theta} \mathbb{E}_{t, U_0, j, k}\left[\mathcal{L}_{GPC}\left(\theta; U_0, \bar{U}_k^{(j)}, \bar{U}_{k-1}^{(j)}, x_k^{(j)}, t\right)\right]$ \label{ae:flow}
    }
\end{algorithm}

We first sample initial states $x_0^{(j)}$ from initial conditions $\mathcal{X}_0$ for $N_E$ parallel simulations. We then perform SPC in each environment, with $N_S$ samples coming from the Gaussian proposal distribution (line \ref{ae:spc_samples}) and the remaining samples from the flow matching policy (line \ref{ae:flow_samples}). The policy samples help improve performance, while the Gaussian samples prevent distribution collapse. After collecting a set of states $x_k^{(j)}$ and action sequences $U_k^{(j)}$, we fit the flow matching model (line \ref{ae:flow}). The expectation is taken over flow timesteps $t \in \mathcal{U}(0, 1)$, initial samples $U_0 \sim \mathcal{N}(0, I)$, parallel environments $j = 1,\dots, N_E$, and simulation steps $k = 1, \dots, K$.



\textbf{GPC benefits from parallelism} throughout Algorithm~\ref{alg:gpc}. In addition to parallel rollouts in the SPC update step (line \ref{ae:rollouts}), we parallelize over simulation environments (line \ref{ae:envs}) and in the model training step (line \ref{ae:flow}). Our implementation leverages the vectorization and parallelization tools in JAX \cite{jax2018github} together with the massively parallel robotics simulation made possible by MuJoCo MJX \cite{mjx}. 

\section{Using a Trained GPC Policy}

\subsubsection{Warm-Starts} In fast feedback loops, the multi-modal expressiveness of generative models presents a challenge: samples at subsequent timesteps can be drawn from different modes, leading to a ``jittering'' (\textit{temporal consistency} \cite{zhao2023learning}) problem. A common solution is to roll out several steps of the action sequence before replanning \cite{chi2023diffusion}. This forces the controller to ``commit'' to a particular mode, but is not suitable for highly dynamic tasks. Other alternatives like action inpainting \cite{black2025real} are effective on quasi-static manipulation tasks, but---as we will show---are not particularly helpful for high-frequency feedback.

We propose a simple alternative inspired by warm-starts in MPC. Rather than starting the flow generation process from $U_0 \sim \mathcal{N}(0, I)$, we start from 
\begin{equation}
    U_0 = (1 - \alpha) \epsilon + \alpha \bar{U}_{k - 1}, \quad \epsilon \sim \mathcal{N}(0, I) 
\end{equation}
where $\alpha \in [0, 1]$ is the \textit{warm-start level}. With $\alpha = 1$, the flow process is started from the previous sample $\bar{U}_{k-1}$, while $\alpha = 0$ recovers the Gaussian proposal distribution. Because the flow matching vector field drives samples toward a mode of the sampling distribution, flows with a high warm-start level $\alpha$ tend to stay close to the same mode as the previous sample, $\bar{U}_{k-1}$. This simple warm-start procedure enables smooth and performant high-frequency control.

\subsubsection{Deploying the Policy} We can use a GPC policy in two ways. A direct application of the policy in receding-horizon fashion, possibly with warm-starts, is simply termed \textbf{\em GPC}. The second \textbf{\em GPC+} strategy uses policy samples to bootstrap SPC, alongside samples from the Gaussian proposal distribution. GPC+ leverages inference-time compute for better performance, but requires a state estimate from which to perform the rollouts. Ordinary GPC does not require a state estimate, as the policy can be conditioned on arbitrary observations.

\subsubsection{Risk-Aware Domain Randomization} Domain randomization (DR) has emerged as a key ingredient for sim-to-real transfer of policies trained in simulation, particularly for reinforcement learning (RL) \cite{handa2023dextreme, tobin2017domain}. 

GPC and massively parallel simulation enable a range of new DR possibilities. In particular, we can modify the SPC rollouts \eqref{eq:spc_rollouts} by simulating each action sequence $U^{(i)}$ in several domains with randomized parameters (e.g., friction coefficients, body masses, etc.). This results in cost values indexed by both sample $i$ and domain $d$, e.g.,
\begin{equation}
    J^{(i, d)} = J(U^{(i)}; x_{k-1}, d).
\end{equation}
We then aggregate this cost data across domains before performing the standard SPC update \eqref{eq:spc_update}.

The simplest choice would be to average over domains,
\begin{equation}\label{eq:average_dr}
    J^{(i)} = \mathbb{E}_d \left[ J^{(i, d)} \right].
\end{equation}
This is analogous to the typical RL domain randomization framework, which considers the expected reward over all domains. But GPC allows for other possibilities as well. We can, for instance, use the worst-case cost,
\begin{equation}
    J^{(i)} = \max_d \left[ J^{(i, d)} \right],
\end{equation} or more sophisticated risk metrics like conditional value-at-risk (CVaR) \cite{rockafellar2000optimization, dixit2023risk}, which takes the expected cost in the $(1 - \beta)$ tail of the distribution:
\begin{equation}\label{eq:cvar_dr}
    J^{(i)} = \inf_{z \in \mathbb{R}} \mathbb{E}_d \left[ z + \frac{\max(J^{(i, d)} - z, 0)}{1 - \beta} \right],
\end{equation}
where $\beta \in [0, 1)$ determines the degree of risk sensitivity. 

These and other risk strategies for online domain randomization are implemented in \texttt{hydrax} \cite{kurtz2024hydrax}, making them readily available for GPC training.

\section{Simulation Studies}\label{sec:experiments}

\begin{figure}
    \centering
    \includegraphics[width=0.13\linewidth]{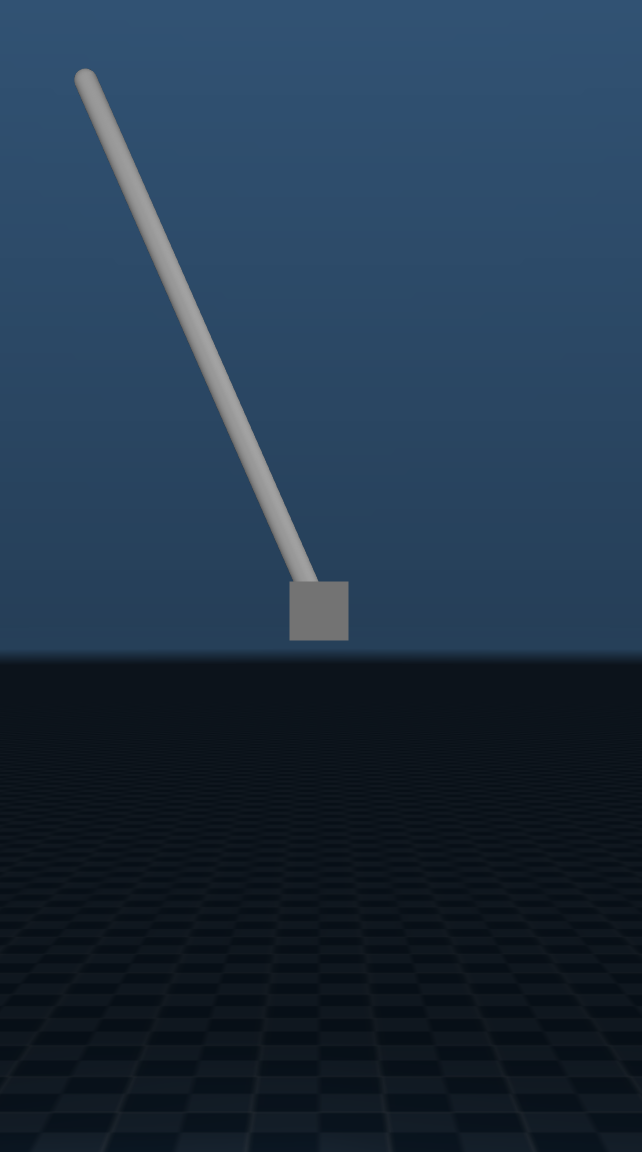}
    \includegraphics[width=0.13\linewidth]{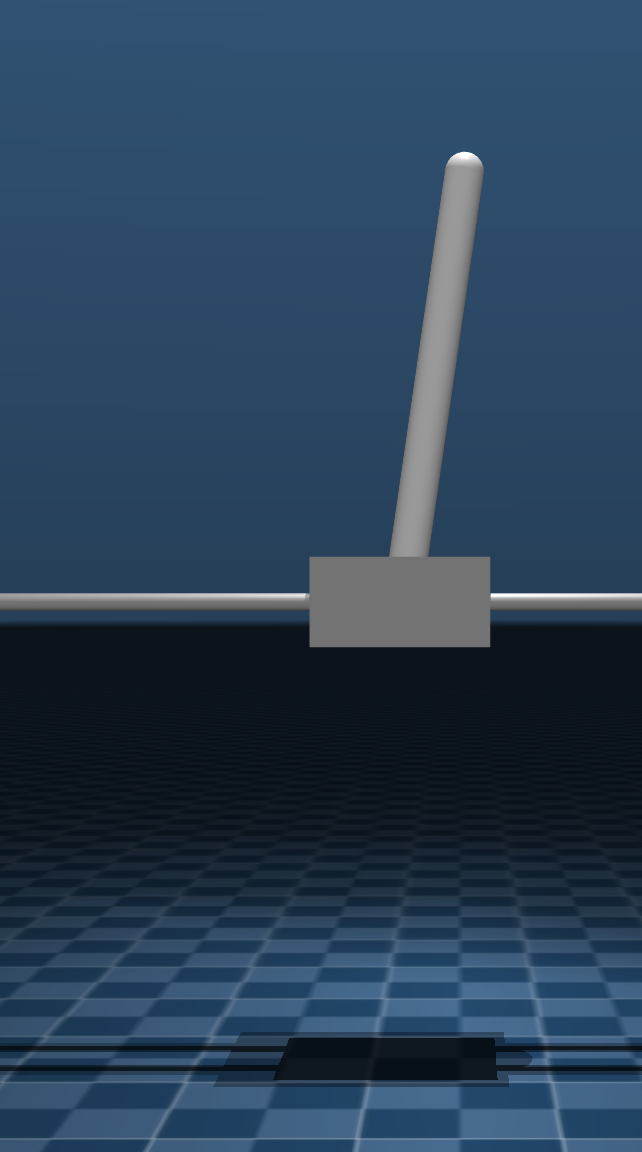}
    \includegraphics[width=0.13\linewidth]{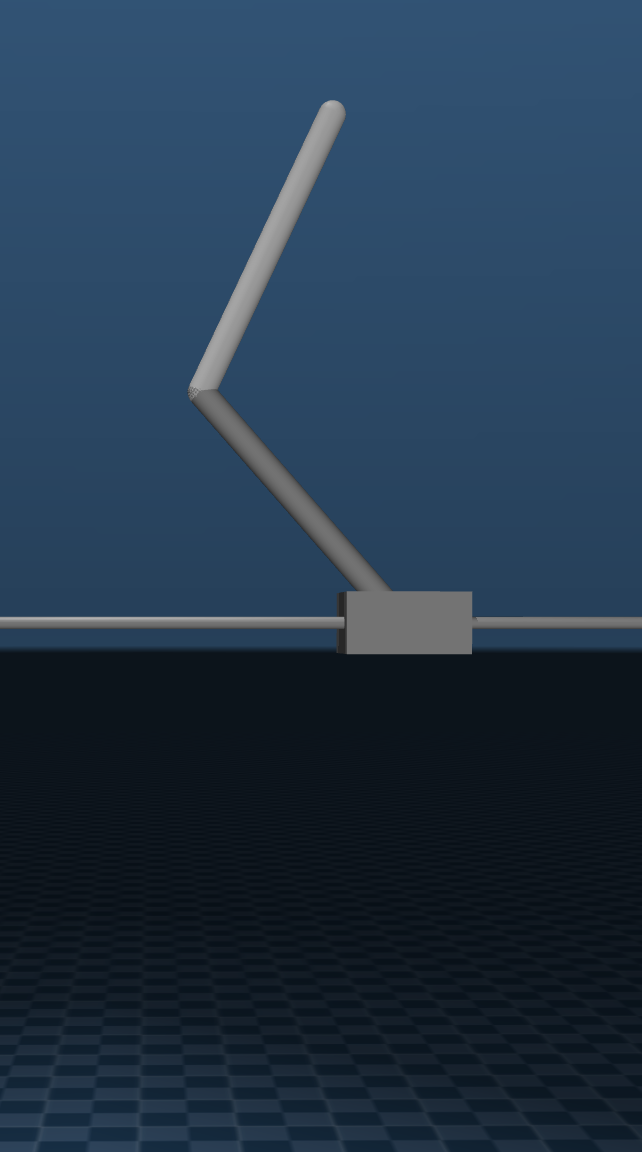}
    \includegraphics[width=0.13\linewidth]{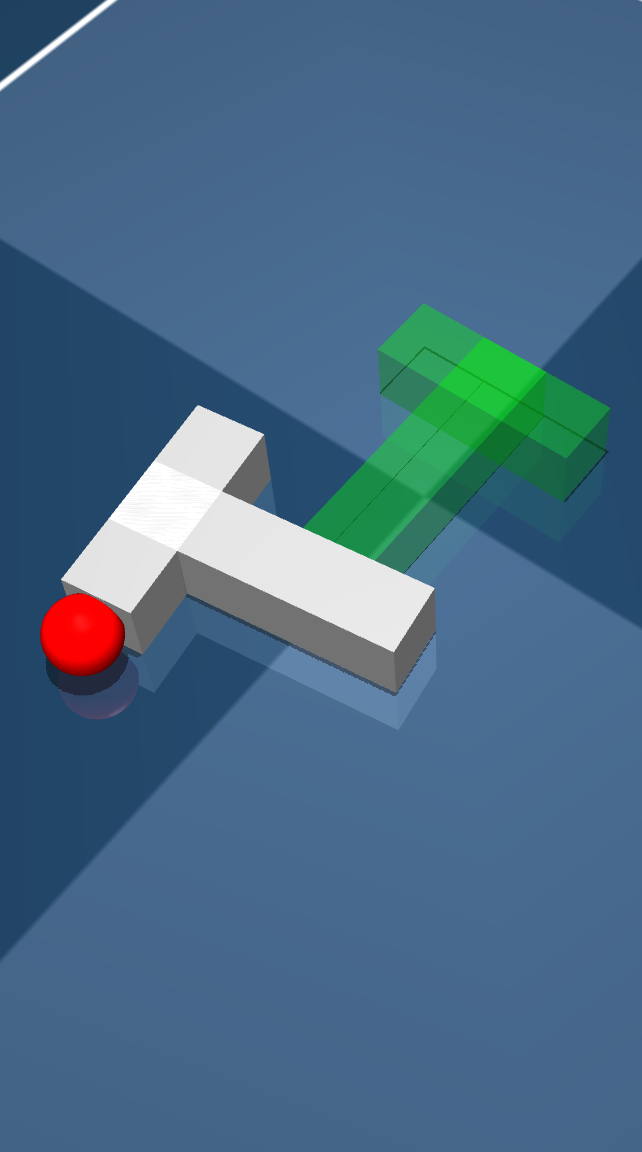}
    \includegraphics[width=0.13\linewidth]{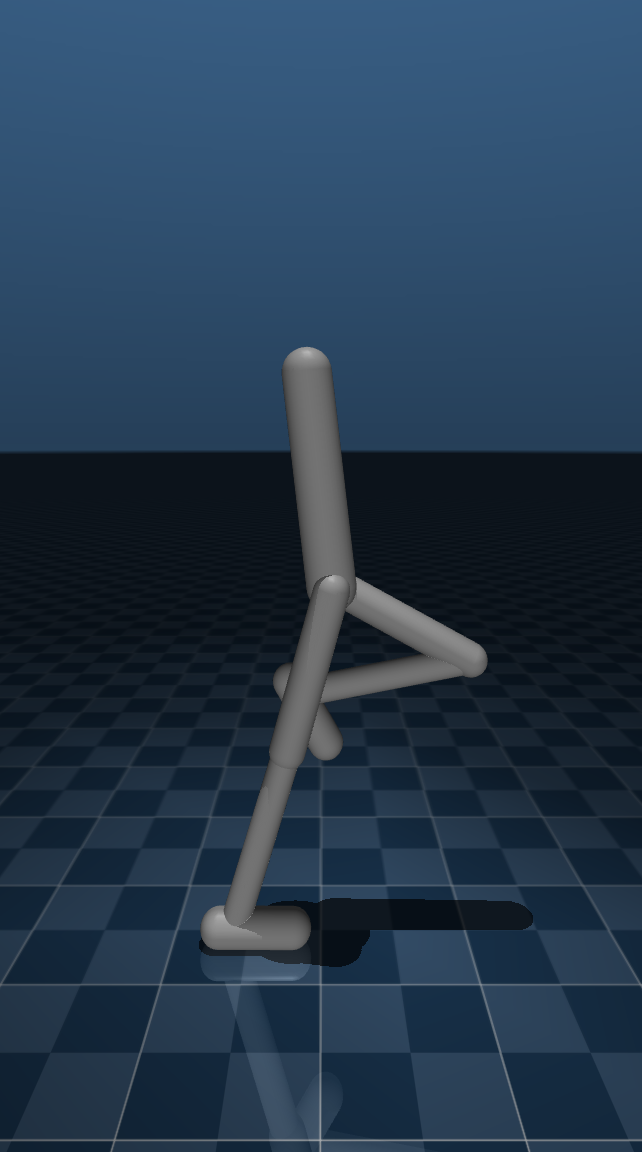}
    \includegraphics[width=0.13\linewidth]{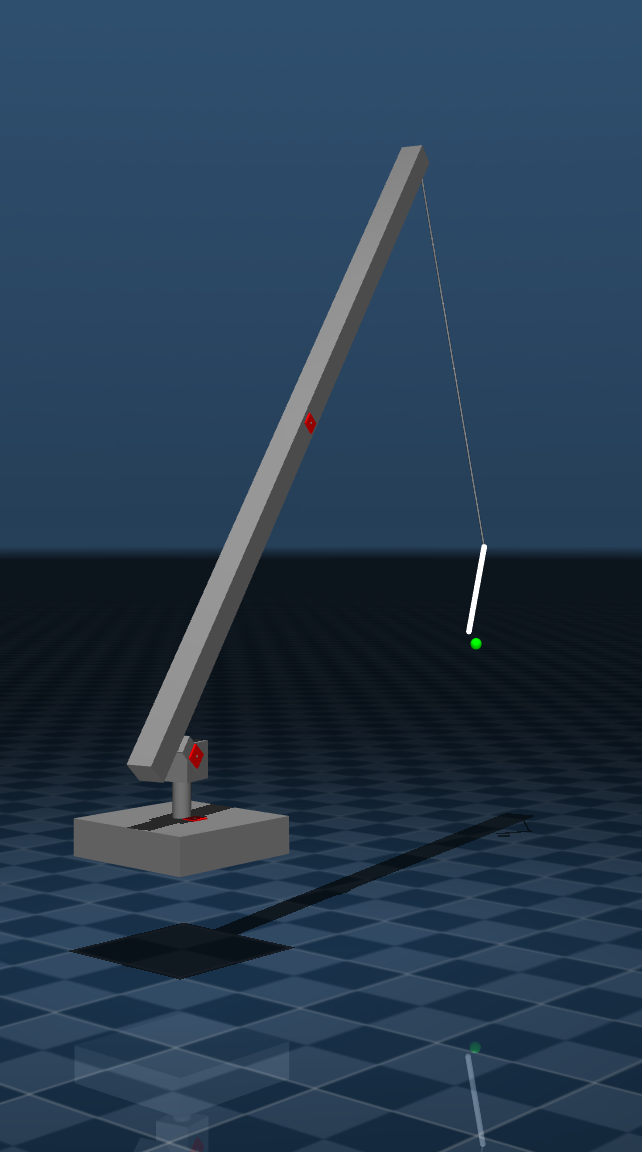}
    \includegraphics[width=0.13\linewidth]{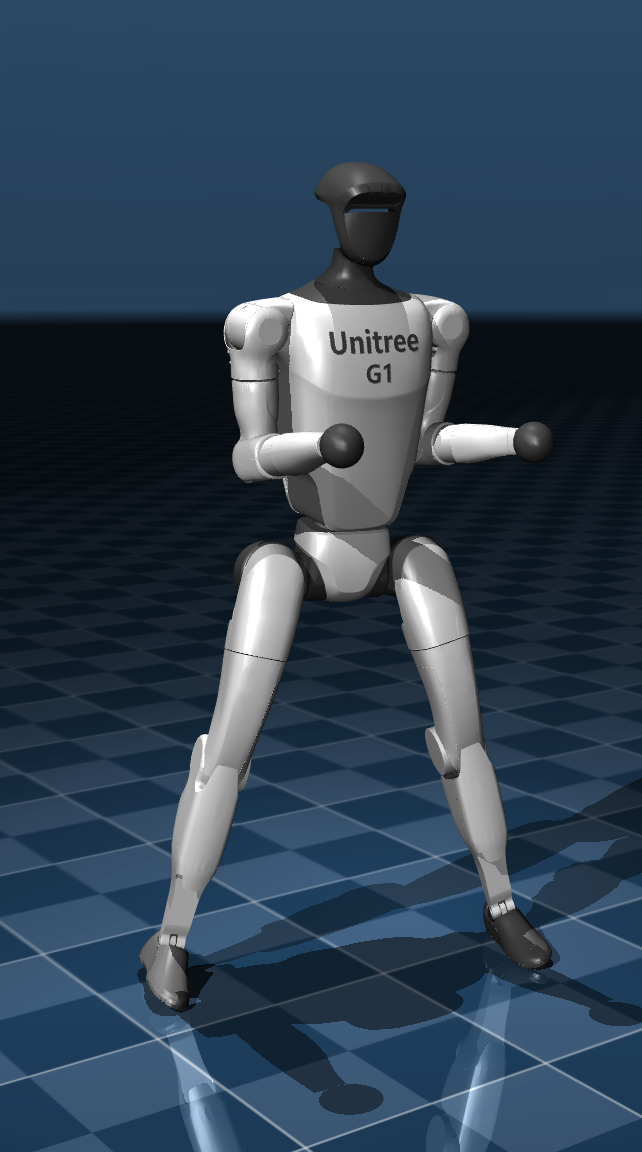}
    \caption{Systems used to evaluate GPC performance in simulation, from left to right: inverted pendulum, cart-pole, double cart-pole, push-T, planar walker, luffing crane, humanoid standup.}
    \label{fig:systems}
\end{figure}

In this section, we aim to answer the following:
\begin{enumerate}
    \item Can GPC perform tasks that require multi-modal reasoning, as well highly dynamic tasks that require high-frequency feedback (Sec.~\ref{sec:policy_performance})?
    \item Does GPC continually improve policy performance over multiple iterations (Sec.~\ref{sec:training_stability})?
    \item How do different domain randomization strategies impact performance (Sec.~\ref{sec:domain_randomization})?
    \item What are the scalability limits of this approach (Sec.~\ref{sec:scalability})?
\end{enumerate}

In short, GPC is effective for systems with fast dynamics at high feedback rates, enjoys the training stability characteristic of supervised learning methods, and enables risk-aware control, but demonstrates scaling limitations on our largest and most difficult example (humanoid standup). We discuss these scalability limits and possible solutions below.

We simulate GPC on seven systems of varying state dimension and task difficulty (see Fig.~\ref{fig:systems}). The \textbf{\em pendulum}, \textbf{\em cart-pole}, and \textbf{\em double cart-pole} are tasked with balancing upright. In \textbf{\em push-T}, an actuated finger pushes a block to a goal pose. The \textbf{\em walker} aims to move forward at a constant velocity, while the \textbf{\em crane} swings its payload to a target position. The \textbf{\em humanoid} attempts to stand up from arbitrary initial configurations. Full details are available with the source code: \url{https://github.com/vincekurtz/gpc}.

The three smallest examples use a multi-layer perceptron for the flow network $v_\theta$, while the others use a convolutional network with FiLM conditioning \cite{perez2018film}, as in \cite{chi2023diffusion}. All simulations ran on a desktop computer with an NVIDIA RTX 4070 (12 GB) GPU. When evaluating trained policies, we use MuJoCo CPU (64-bit) rather than MJX (32-bit), resulting in a small sim-to-sim gap. 

\subsection{Policy Performance}\label{sec:policy_performance}

\begin{figure*}
    \centering
    \begin{subfigure}{0.95\linewidth}
        \centering
        \includegraphics[width=\linewidth]{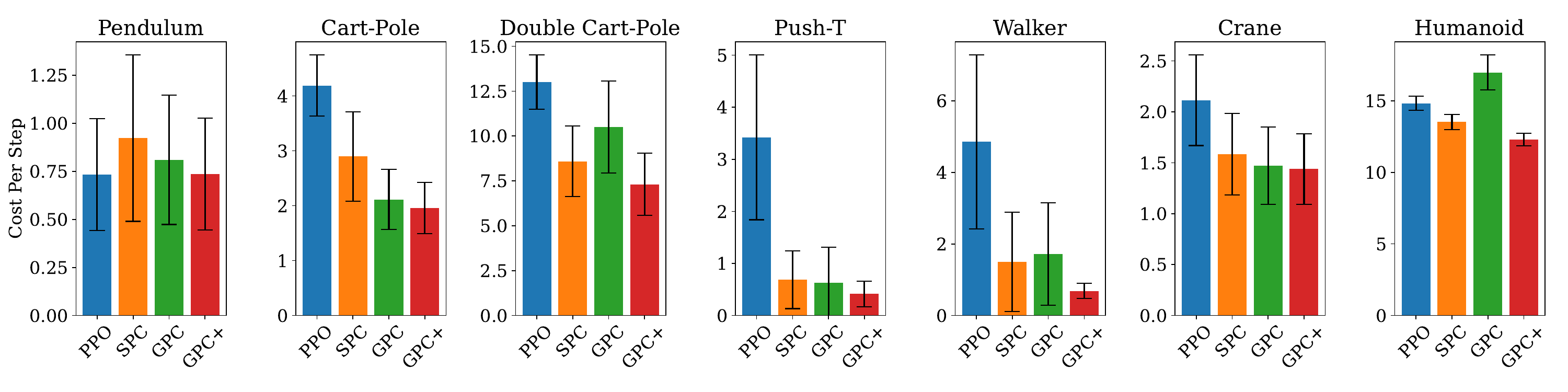}
        \caption{Comparing RL (PPO), predictive sampling (SPC), and our methods (GPC, GPC+). Applying the GPC policy directly provides performance on-par-with or better-than SPC in all cases except humanoid standup. GPC+ meets or exceeds the performance of the other methods across all examples.}
        \label{fig:simulation_costs:baselines}
    \end{subfigure}
    \begin{subfigure}{0.95\linewidth}
        \centering
        \includegraphics[width=\linewidth]{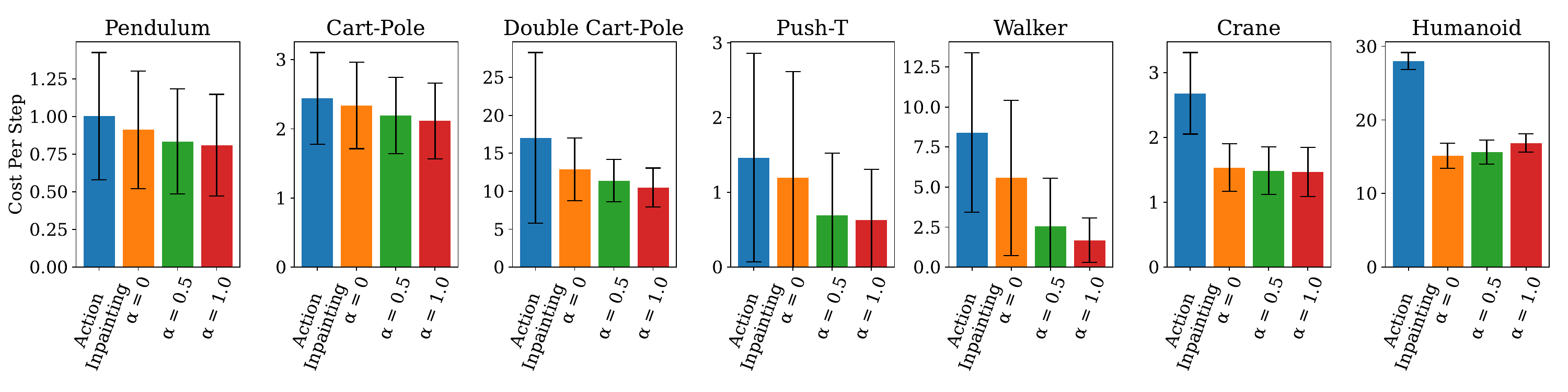}
        \caption{Comparing warm-start strategies: action inpainting \cite{black2025real}, no warm-start ($\alpha = 0$), partial warm-start ($\alpha = 0.5$), and full warm-start $\alpha = 1.0$.}
        \label{fig:simulation_costs:warm_starts}
    \end{subfigure}
    \caption{Performance comparisons showing average cost per time step (lower is better). Black bars indicate standard deviation over 100 ten-second simulations from randomized initial conditions. }
    \label{fig:simulation_costs}
\end{figure*}

Footage of closed-loop GPC performance on each of the examples is shown in the accompanying video (\url{https://youtu.be/mjL7CF877Ow}). To evaluate performance quantitatively, we perform 100 randomly-initialized simulations and record the average cost per time step as a performance metric. Figure~\ref{fig:simulation_costs:baselines} compares our approach (GPC, GPC+) with PPO and SPC baselines. For PPO training, we use the same cost (negative reward), network size, batch size, and total number of simulation time steps\footnote{Full training details can be found in the open-source implementation. While it is possible that other hyperparameters settings exist that could produce better PPO performance, we took considerable effort via manual hyperparameter tuning to ensure PPO produced as competitive of performance as possible.} as GPC. SPC and GPC+ both use $N = 128$ rollouts. Interestingly, \textbf{GPC and GPC+ perform on par or better than PPO with the same amount of training data}. Of course, further hyperparameter and reward tuning could improve both PPO and GPC performance: a systematic hyperparameter sensitivity comparison is left for future work.

Figure~\ref{fig:simulation_costs:warm_starts} compares our warm-start strategy with action inpainting \cite{black2025real}, both using the same GPC policy. For action inpainting, we use the soft-masking pseudoinverse guidance strategy of \cite[Eq. 2-4]{black2025real}. Interestingly, action inpainting---a state-of-the-art method for temporal consistency enforcement in behavior cloning---degrades performance on these high-frequency tasks. This is likely because action inpainting is designed for relatively slow, quasi-static tasks with significant inference delays. In contrast, GPC inference times range between 1 and 10 milliseconds, resulting in feedback rates between 100-1000 Hz.

Note that Fig.~\ref{fig:simulation_costs} provides some insight into relative performance, but is limited. For instance, on the walker we find that GPC enables smoother actions with far less ``stumbling'' than SPC, though cost per step indicates similar performance.

GPC can handle multi-modal action distributions, as evidenced by effectiveness on the push-T task, which requires multi-modal reasoning to reach around the block \cite{chi2023diffusion}. Interestingly, GPC training takes under 20 minutes, while training a similar diffusion policy takes around an hour, not counting the time to gather demonstrations \cite{chi2023diffusion}.

\begin{figure}
    \centering
    \includegraphics[width=0.9\linewidth]{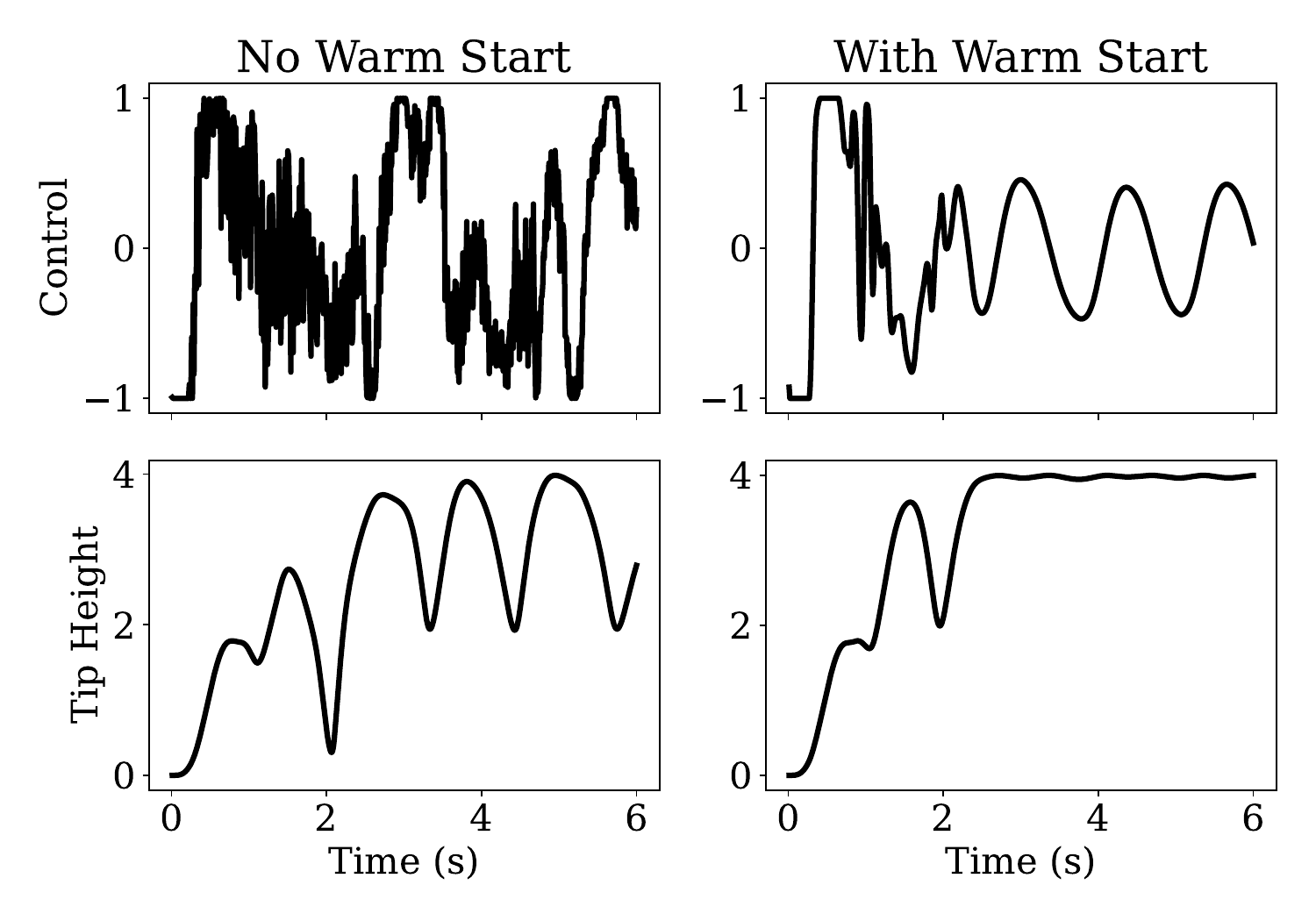}
    \caption{Closed-loop double cart-pole performance with and without warm-starts. The warm-started policy (right) produces smooth actions and can successfully balance. Without warm-starts (left), actions jitter between modes, and the robot fails to balance.}
    \label{fig:warm-starts}
\end{figure}

More importantly, \textbf{GPC can control systems with fast dynamics at high control rates}. The double cart-pole illustrates this fact, as well as the importance of warm-starts. Fig.~\ref{fig:warm-starts} shows performance with and without warm-starts. Without warm-starts (left, $\alpha = 0$), the control actions are dominated by significant noise (top plot) and the system cannot swing upright (bottom plot). Warm-starts (right, $\alpha = 1$), lead to smoother control actions, and the robot successfully balances around the upright configuration. The controller can respond rapidly to complex system dynamics, as evidenced by the rapid changes between 1 and 2 seconds in Fig. \ref{fig:warm-starts}.

\subsection{Training Stability}\label{sec:training_stability}

\begin{figure*}
    \centering
    \includegraphics[width=\linewidth]{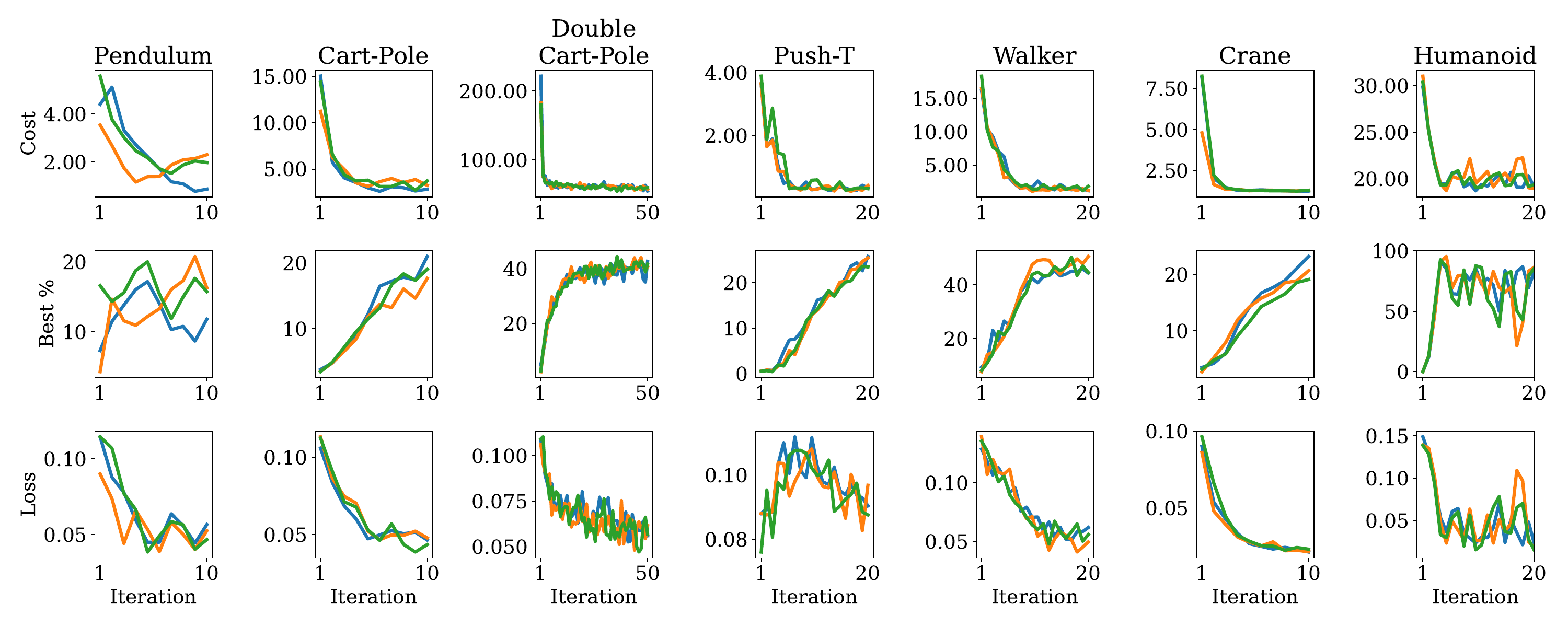}
    \caption{Training curves showing the average cost $J$, percent of states in which the flow-matching policy generated the best action sequence, and the loss $\mathcal{L}_{GPC}$ from three random seeds. GPC is able to leverage the training stability of supervised learning while avoiding the need for demonstrations.}
    \label{fig:training_curves}
\end{figure*}

During the training process, we find that the average cost of policy samples decreases monotonically between iterations, modulo noise from initial conditions. This indicates that GPC's cycle of training and sampling continually improves performance. These and other training curves are shown in Fig.~\ref{fig:training_curves}. While we leave a systematic hyperparameter sensitivity study for future work, we empirically observe that \textbf{GPC benefits from the training stability of supervised learning}. This contrasts with reinforcement learning methods, which can exhibit high sensitivity to reward tuning, implementation details, and even the random seed used for training \cite{engstrom2019implementation, andrychowicz2020matters}. 

\subsection{Risk-Aware Domain Randomization}\label{sec:domain_randomization}

\begin{table}
    \centering
    \begin{tabular}{|c|ccc|}
        \hline
        & No DR & Average DR & CVaR DR \\
        \hline
         No model error & 106 & \textbf{103} & 133   \\
        \hline
         With model error & 165 & 184 & \textbf{139} \\
        \hline
    \end{tabular}
    \caption{Time (seconds) for the crane to visit a sequence of 50 randomly generated payload targets, under policies trained without DR, with standard DR \eqref{eq:average_dr}, and with a risk-averse CVaR strategy \eqref{eq:cvar_dr}. CVaR improves robustness at the cost of worse performance under nominal conditions.}
    \label{tab:crane_dr}
\end{table}

We use the luffing crane example to explore the impact of different DR strategies. Specifically, we train three GPC policies: one with no DR, one with standard average-cost DR \eqref{eq:average_dr}, and one with a more conservative CVaR \eqref{eq:cvar_dr} strategy ($\beta = 0.25$). We use 8 randomized domains, each with slightly different joint damping, payload mass, payload inertia, and actuator gains.

After training, we apply GPC with warm-starts. To evaluate closed-loop performance, we randomize 50 target locations for the payload to visit. The target moves to the next location once the payload lies within 15 cm or after 10 seconds, whichever comes first. Table~\ref{tab:crane_dr} reports the total time to visit all 50 targets: lower times are better. The same target location sequence is used for each policy.

Without model error in the simulator, both the non-randomized policy and average-case DR perform significantly better than the more conservative CVaR policy. When we add model error to the simulator (lower joint damping, heavier payload mass), all methods perform worse. But the more conservative CVaR policy degrades the least, and significantly outperforms the others.

\subsection{Scalability}\label{sec:scalability}

We chose example systems possessing 1 to 29 degrees-of-freedom, in order to assess the scalability of Algorithm~\ref{alg:gpc}. This \textbf{scalability limit is reached with the largest humanoid example}: the GPC policy alone is unable to reliably stand up, though GPC+ remains effective. Further cost and hyperparameter tuning, curriculum training, and more computation could likely improve performance. 

\section{Limitations and Future Work}\label{sec:limitations}

Limited effectiveness of basic GPC on the humanoid standup example is the most severe limitation of our method, as noted above. We believe that \textbf{value function learning will be a key advance in overcoming this limitation}. Besides being a key element of state-of-the-art reinforcement learning methods, value learning would enable a reduced planning horizon $T$ in Problem \ref{eq:ocp}. Reducing the planning horizon reduces the dimensionality of the sampling space, making planning easier while maintaining long-horizon reasoning. Methods that leverage connections between the gradient of the value function and the flow field $v_\theta$---which is in turn closely related to the score $\nabla \log p(U \mid x)$ and therefore the gradient of the cost \eqref{eq:ocp_compact}---are of particular interest.

For the basic GPC framework introduced here, we run $N$ short simulations---and even more under a risk-aware DR strategy---to generate a single training data point. While these simulation rollouts are significantly faster and cheaper to collect than human demonstrations, methods that more fully use \textit{all} of the data from SPC rollouts could further improve the sample efficiency of GPC.

Other performance improvements could come from more benign algorithmic details. For instance, we represent action sequences with simple zero-order-hold splines. Higher-order splines \cite{howell2022predictive} or alternative parameterizations could be more effective. Better choices of action space, such as using task-space/end-effector coordinates rather than joint coordinates, could also be useful. Actuation limits, which are a critical component of many robotics tasks, are not handled in any particularly special way. Leveraging recent advances in constrained generative modeling \cite{fishman2024diffusion, fishman2024metropolis,kurtz2024equality} to do so is another potentially fruitful area for future work. 

Hardware experiments will provide an important platform for exploring policies that are conditioned on complex observations like raw sensor data, images, or foundation model embeddings \cite{oquab2023dinov2}. Integrated simulation and rendering in the recently-released MuJoCo playground \cite{mujoco_playground_2025} could provide a useful platform for training image-conditioned policies. 

\section{Conclusion}

We introduced generative predictive control (GPC), a framework for learning flow matching policies on dynamic tasks that are easy to simulate but difficult to demonstrate. GPC leverages tight connections between generative modeling and sampling-based predictive control to generate training data for supervised learning without expert demonstrations. We showed how warm-started GPC policies enable real-time high-frequency control, ensuring temporal consistency via warm-starts. GPC may offer a path toward including dynamic and difficult-to-demonstrate tasks in a generalist policy or large behavior model that combines data from many tasks \cite{black2024pi_0, lbm, team2025gemini}. Future work will focus on validating the GPC framework on hardware, incorporating value function learning, and training multi-task policies.

\balance
\bibliographystyle{unsrt}
\bibliography{references}

\include{appendix}

\end{document}

%% file: packages.tex
\usepackage{cite}
\usepackage{amsmath,amssymb,amsfonts}
\usepackage{mathtools}
\usepackage{bm}
\usepackage{graphicx}
\usepackage{textcomp}
\usepackage{xcolor,color}

\usepackage{amsthm}
\usepackage{amsmath}
\usepackage{tikz}
\usetikzlibrary{positioning,automata}
\usepackage{subcaption}
\usepackage{listings}
\usepackage{pifont} 
\usepackage[linesnumbered,ruled,lined]{algorithm2e}
\usepackage{hyperref}
\usepackage{balance}

%% file: notation.tex
\newtheorem{proposition}{Proposition}
\newtheorem{remark}{Remark}

\definecolor{codegreen}{rgb}{0,0.6,0}
\definecolor{codegray}{rgb}{0.5,0.5,0.5}
\definecolor{codepurple}{rgb}{0.58,0,0.82}
\definecolor{backcolour}{rgb}{0.95,0.95,0.92}

\lstdefinestyle{mystyle}{
    backgroundcolor=\color{backcolour},   
    commentstyle=\color{codegreen},
    keywordstyle=\color{magenta},
    numberstyle=\tiny\color{codegray},
    stringstyle=\color{codepurple},
    basicstyle=\ttfamily\footnotesize,
    breakatwhitespace=false,         
    breaklines=true,                 
    captionpos=b,                    
    keepspaces=true,                 
    numbersep=5pt,                  
    showspaces=false,                
    showstringspaces=false,
    showtabs=false,                  
    tabsize=2
}
\SetKwComment{Comment}{// }{}

\newcommand{\x}{\mathbf{x}}
\newcommand{\y}{\mathbf{y}}



%% file: appendix.tex
\clearpage

\appendices

\section{Cosine-Distance Objective Weighting}\label{apx:cosine_similarity}

We use a small modification of the flow matching loss based on the cosine similarity score
\begin{equation}
 S_C(\x, \y) = \frac{\x \cdot \y}{\|\x\|\|\y\|},
\end{equation}
which measures directional similarity between vectors $\x$ and $\y$. $S_C$ outputs a scalar in $[-1, 1]$, with high values indicating a high degree of similarity. 

In our case, we are more confident that the flow-matching objective $\mathcal{L}_{GPC}$ \eqref{eq:flow_objective} drives samples toward the true (but unknown) optimal action sequence $U^*$ if the update $\bar{U}_k - \bar{U}_{k-1}$ and the flow target $\bar{U}_k - U_0$ are close. This is illustrated in the ``flow matching'' portion of Fig.~\ref{fig:hero}. Darker shading indicates values of $U_0$ (typically drawn from a standard Gaussian) that are more informative. In particular, samples in the upper left-hand corner could be pushed toward $\bar{U}_k$ but away from $U^*$, while samples in the lower right would flow in the correct direction.

To capture this mathematically,  we define weights
\begin{multline}\label{eq:weights}
 w(\bar{U}_k, \bar{U}_{k-1}, U_0) = \\
 \exp\left(-\gamma (1 - S_C(\bar{U}_k - \bar{U}_{k-1}, \bar{U}_k - U_0)))\right),
\end{multline}
where the hyperparameter $\gamma > 0$ defines a decay rate as the two vectors move further apart (our implementation uses $\gamma = 2.0$). This gives us $w = 1$ if $\bar{U}_{k-1}$ and $U_0$ are close in the cosine similarity sense, and $w \to 0$ as they move further apart. The final training loss is given by
\begin{equation}
 w(\bar{U}_k, \bar{U}_{k-1}, U_0) \mathcal{L}_{GPC}(\theta; U_0; \bar{U}_k, x_k, t).
\end{equation}

This modification is motivated by the fact that while the SPC update is a monte-carlo score ascent step, it does not obtain a true sample $U^*$ from the optimal target distribution. In other words, $\bar{U}_{k}$ is closer to an optimal sample than $\bar{U}_{k-1}$, but it is not the case that $\bar{U}_k = U^*$. This is an important distinction between the GPC setting and the typical generative modeling setting, where samples from the target distribution are freely available. 

\begin{table*}
    \small
    \centering
    \begin{tabular}{c|ccccccc}
         & Pendulum & Cart-Pole & \begin{tabular}{c}Double\\Cart-Pole\end{tabular} & Push-T & Walker & Crane & Humanoid \\
         \hline
         DoFs & 1 & 2 & 3 & 5 & 9 & 7 & 29 \\
         Num. actuators & 1 & 1 & 1 & 2 & 6 & 3 & 23 \\
         Observation Size & 3 & 5 & 8 & 5 & 18 & 23 & 56 \\
         Planning horizon (s) & 0.5 & 1.0 & 0.8 & 0.5 & 0.6 & 0.8 & 0.9 \\
         Planning knots & 5 & 10 & 10 & 5 & 4 & 2 & 3\\
    \end{tabular}
    \caption{Example task specification parameters. }
    \label{tab:example_details}
\end{table*}

\section{Example System Details}\label{apx:example_details}

Details regarding each of the example systems from Section~\ref{sec:experiments} can be found below, with further details in Table~\ref{tab:example_details}. We represent the action sequence $U$ with a zero-order-hold spline over a fixed number of knot points. Using few knot points is critical for good SPC performance, particularly on the higher-DoF examples.

\textbf{Inverted pendulum:} This simple one-dimensional system requires swinging a pendulum to the upright position and balancing it there. Torque limits prevent the pendulum from swinging directly upright: the policy must gradually pump energy into the system. The observation $y = h(x)$ consists of sine and cosine of the angle, along with angular velocity. 

\textbf{Cart-pole:} An unactuated pendulum is mounted on an actuated cart. Control actions are torques applied to the cart, and the task is to balance the pendulum upright. While this is a relatively simple nonlinear system, obtaining successful teleoperated demonstrations would be difficult. The observations are sine and cosine of the pendulum angle, position of the cart, and linear and angular velocities.

\textbf{Double cart-pole:} In this extension of the cart-pole example, an unactuated double pendulum is mounted on the cart. The fast and chaotic double pendulum dynamics make this task particularly challenging. As for the cart-pole, observations include sine and cosine of pendulum angles, cart position, and linear and angular velocities.

\textbf{Push-T:} A robotic finger pushes a T-shaped block to a goal position and orientation on a table. This task has been solved with behavior cloning \cite{chi2023diffusion}, and is a standard example of a task that requires multi-modal reasoning. The observations include pusher position, block position, and block orientation.

\textbf{Planar biped:} A robot walker, constrained to the sagittal plane, is tasked with moving forward at 1.5 m/s. The high dimensionality of this system would make teleoperation for behavior cloning difficult. Successful locomotion also requires fast replanning. The observation is the full system state (positions and velocities), excluding the horizontal position.

\textbf{Luffing crane:} A swinging payload is attached via a rope to a luffing crane. Control actions are target boom angles and rope length. This underactuated system provides a particularly useful testbed for investigating the impact of modeling errors and domain randomization strategies. The observation includes crane joint angles and the position and velocity of the payload relative to the target.

\textbf{Humanoid standup:} A Unitree G1 humanoid model is tasked with reaching a standing configuration. Initial conditions are random joint angles, joint velocities, and base orientation, so that the robot begins sprawled on the ground. The observation includes all joint angles and joint velocities, floating base velocity, floating base position, and floating base orientation relative to the upright.

\section{Training Details}\label{apx:training}

Hyperparameters for policy training and inference are summarized in Table~\ref{tab:summary}. Full details can be found in the open-source implementation. When evaluating a trained SPC or GPC+ policy, we used a total of 128 simulation rollouts. For SPC, all of these are drawn from a Gaussian proposal distribution. For GPC+, 64 are from the Gaussian distribution and 64 are from the flow matching policy.

\begin{table*}[]
    \small
    \centering
    \begin{tabular}{c|ccccccc}
         & Pendulum & Cart-Pole & \begin{tabular}{c}Double\\Cart-Pole\end{tabular} & Push-T & Walker & Crane & Humanoid \\
         \hline
         Architecture & MLP & MLP & MLP & CNN & CNN & CNN & CNN \\
         Iterations & 10 & 10 & 50 & 20 & 20 & 10 & 20 \\
         Train (min:sec) & 00:30 & 01:20 & 17:10 & 17:30 & 9:36 & 6:15 & 113:18 \\
         \hline
         Envs. ($N_E$) & 128 & 128 & 256 & 128 & 128 & 512 & 128 \\
         SPC samp. ($N_S$) & 8 & 8 & 16 & 128 & 16 & 4 & 32 \\
         Policy samp. ($N_P$) & 2 & 2 & 16 & 32 & 16 & 2 & 32 \\
         Episode len. (sec) & 4.0 & 2.0 & 4.0 & 4.0 & 5.0 & 5.0 & 8.0 \\
         \hline
         Num. Params. & 1541 & 5898 & 20362 & 19990 & 49768 & 48680 & 368567 \\
         Batch size & 128 & 128 & 128 & 128 & 128 & 1024 & 128\\
         Learning rate & $10^{-3}$ & $10^{-3}$ & $10^{-3}$ & $10^{-3}$ & $10^{-3}$ & $10^{-3}$ & $10^{-3}$\\
         Epochs & 10 & 100 & 10 & 10 & 10 & 20 & 10  \\
         \hline
         Inference (ms) & 1.0 & 1.3 & 1.3 & 3.8 & 8.1 & 2.4 & 4.0 \\
         ODE step size $\delta t$ & 0.1 & 0.1 & 0.1 & 0.1 & 0.1 & 0.1 & 0.1 \\
         Ctrl. Freq. (Hz) & 50 & 50 & 50 & 50 & 50 & 30 & 50 \\
    \end{tabular}
    \caption{Summary of training and inference hyperparameters. }
    \label{tab:summary}
\end{table*}

The hyperparameters in Table~\ref{tab:summary} are not highly optimized: our primary objective is merely to confirm the ability of GPC to train usable generative policies. We leave a systematic hyperparameter sensitivity study for future work.

%% file: references.bib
@article{fishman2024metropolis,
  title={Metropolis sampling for constrained diffusion models},
  author={Fishman, N. and Klarner, L. and Mathieu, E. and Hutchinson, M. and De Bortoli, V.},
  journal={Advances in Neural Information Processing Systems},
  volume={36},
  year={2024}
}

@article{fishman2024diffusion,
  title={Diffusion Models for Constrained Domains},
  author={Fishman, N. and Klarner, L. and De Bortoli, V. and Mathieu, E. and Hutchinson, M.J},
  journal={Trans. Machine Learning Research},
  year={2024}
}

@article{posa2014direct,
  title={A direct method for trajectory optimization of rigid bodies through contact},
  author={Posa, Michael and Cantu, Cecilia and Tedrake, Russ},
  journal={The International Journal of Robotics Research},
  volume={33},
  number={1},
  pages={69--81},
  year={2014},
  publisher={Sage Publications Sage UK: London, England}
}

@article{kurtz2023inverse,
  title={Inverse dynamics trajectory optimization for contact-implicit model predictive control},
  author={Kurtz, Vince and Castro, Alejandro and {\"O}nol, Aykut {\"O}zg{\"u}n and Lin, Hai},
  journal={  arXiv:2309.01813},
  year={2023}
}

@article{song2019generative,
  title={Generative modeling by estimating gradients of the data distribution},
  author={Song, Y. and Ermon, S.},
  journal={NeurIps},
  volume={32},
  year={2019}
}

@article{chi2023diffusion,
  title={Diffusion policy: Visuomotor policy learning via action diffusion},
  author={Chi, C. and Feng, S. and Du, Y. and Xu, Z. and Cousineau, E. and Burchfiel, B. and Song, S.},
  journal={  arXiv:2303.04137},
  year={2023}
}

@article{pan2024model,
  title={Model-Based Diffusion for Trajectory Optimization},
  author={Pan, Chaoyi and Yi, Zeji and Shi, Guanya and Qu, Guannan},
  journal={  arXiv:2407.01573},
  year={2024}
}

@article{xue2024full,
  title={Full-Order Sampling-Based MPC for Torque-Level Locomotion Control via Diffusion-Style Annealing},
  author={Xue, Haoru and Pan, Chaoyi and Yi, Zeji and Qu, Guannan and Shi, Guanya},
  journal={  arXiv:2409.15610},
  year={2024}
}

@misc{jax2018github,
  author = {James Bradbury and Roy Frostig and Peter Hawkins and Matthew James Johnson and Chris Leary and Dougal Maclaurin and George Necula and Adam Paszke and Jake Vander{P}las and Skye Wanderman-{M}ilne and Qiao Zhang},
  title = {{JAX}: composable transformations of {P}ython+{N}um{P}y programs},
  url = {http://github.com/jax-ml/jax},
  version = {0.3.13},
  year = {2018},
}

@article{song2020score,
  title={Score-based generative modeling through stochastic differential equations},
  author={Song, Y. and Sohl-Dickstein, J. and Kingma, D.P. and Kumar, A. and Ermon, S. and Poole, B.},
  journal={arXiv:2011.13456},
  year={2020}
}

@article{williams2017model,
  title={Model predictive path integral control: From theory to parallel computation},
  author={Williams, G. and Aldrich, A. and Theodorou, E.A.},
  journal={J. Guidance, Control, and Dynamics},
  volume={40},
  number={2},
  pages={344--357},
  year={2017}
}

@article{black2024pi_0,
  title={$\pi_0$: A Vision-Language-Action Flow Model for General Robot Control},
  author={Black, K. and Brown, N. and Driess, D. and Esmail, A. and Equi, M. and Finn, C. and Fusai, N. and Groom, L. and Hausman, K. and Ichter, B. et. al.},
  journal={arXiv:2410.24164},
  year={2024}
}

@inproceedings{fu2024mobile,
  title={Mobile ALOHA: Learning Bimanual Mobile Manipulation using Low-Cost Whole-Body Teleoperation},
  author={Fu, Z. and Zhao, T.Z. and Finn, C.},
  booktitle={Conf. on Robot Learning}
}

@article{zhao2023learning,
  title={Learning fine-grained bimanual manipulation with low-cost hardware},
  author={Zhao, Tony Z and Kumar, Vikash and Levine, Sergey and Finn, Chelsea},
  journal={  arXiv:2304.13705},
  year={2023}
}

@article{aydinoglu2024consensus,
  title={Consensus complementarity control for multi-contact mpc},
  author={Aydinoglu, Alp and Wei, Adam and Huang, Wei-Cheng and Posa, Michael},
  journal={IEEE Transactions on Robotics},
  year={2024},
  publisher={IEEE}
}

@article{howell2022predictive,
  title={Predictive sampling: Real-time behaviour synthesis with mujoco},
  author={Howell, Taylor and Gileadi, Nimrod and Tunyasuvunakool, Saran and Zakka, Kevin and Erez, Tom and Tassa, Yuval},
  journal={  arXiv:2212.00541},
  year={2022}
}

@inproceedings{williams2016aggressive,
  title={Aggressive driving with model predictive path integral control},
  author={Williams, G. and Drews, P. and Goldfain, B. and Rehg, J.M. and Theodorou, E.A.},
  booktitle={IEEE Int. Conf. Robotics and Automation},
  pages={1433--1440},
  year={2016}
}

@article{li2024drop,
  title={DROP: Dexterous Reorientation via Online Planning},
  author={Li, A.H. and Culbertson, P. and Kurtz, V. and Ames, A.D.},
  journal={  arXiv:2409.14562},
  year={2024}
}

@article{rubinstein1999cross,
  title={The cross-entropy method for combinatorial and continuous optimization},
  author={Rubinstein, Reuven},
  journal={Methodology and computing in applied probability},
  volume={1},
  pages={127--190},
  year={1999},
  publisher={Springer}
}

@article{makoviychuk2021isaac,
  title={Isaac gym: High performance gpu-based physics simulation for robot learning},
  author={Makoviychuk, Viktor and Wawrzyniak, Lukasz and Guo, Yunrong and Lu, Michelle and Storey, Kier and Macklin, Miles and Hoeller, David and Rudin, Nikita and Allshire, Arthur and Handa, Ankur and others},
  journal={  arXiv:2108.10470},
  year={2021}
}

@misc{mjx,
  title={MuJoCo XLA (MJX)},
  author={MuJoCo~XLA~Authors},
  note={\url{https://mujoco.readthedocs.io/en/stable/mjx.html}},
  year={2025},
}

@software{Genesis,
  author = {Genesis Authors},
  title = {Genesis: A Universal and Generative Physics Engine for Robotics and Beyond},
  month = {December},
  year = {2024},
  url = {https://github.com/Genesis-Embodied-AI/Genesis}
}

@misc{kurtz2024hydrax,
  title={Hydrax: Sampling-based model predictive control on GPU with JAX and MuJoCo MJX},
  author={Kurtz, Vince},
  year={2024},
  url={https://github.com/vincekurtz/hydrax}
}

@article{lipman2022flow,
  title={Flow matching for generative modeling},
  author={Lipman, Y. and Chen, R.T.Q and Ben-Hamu, H. and Nickel, M. and Le, M.},
  journal={arXiv:2210.02747},
  year={2022}
}

@article{lbm,
  title={A careful examination of large behavior models for multitask dexterous manipulation},
  author={TRI LBM Team},
  journal={arXiv:2507.05331},
  year={2025}
}

@misc{mujoco_playground_2025,
  title = {MuJoCo Playground: An open-source framework for GPU-accelerated robot learning and sim-to-real transfer.},
  author = {Zakka, K. and Tabanpour, B. and Liao, Q. and Haiderbhai, M. et. al. },
  year = {2025},
  publisher = {GitHub},
  url = {https://github.com/google-deepmind/mujoco_playground}
}

@article{oquab2023dinov2,
  title={Dinov2: Learning robust visual features without supervision},
  author={Oquab, M. and Darcet, T. and Moutakanni, T. and Vo, H. and Szafraniec, M. and Khalidov, V. and Fernandez, P. and Haziza, D. and Massa, F. and El-Nouby, A. et. al.},
  journal={  arXiv:2304.07193},
  year={2023}
}

@article{kurtz2024equality,
  title={Equality Constrained Diffusion for Direct Trajectory Optimization},
  author={Kurtz, Vince and Burdick, Joel W},
  journal={  arXiv:2410.01939},
  year={2024}
}

@inproceedings{andrychowicz2020matters,
  title={What matters for on-policy deep actor-critic methods? a large-scale study},
  author={Andrychowicz, M. and Raichuk, A. and Sta{\'n}czyk, P. and Orsini, M. and Girgin, S. and Marinier, R. and Hussenot, L. and Geist, M. and Pietquin, O. and Michalski, M. et. al},
  booktitle={Int. Conf. Learning Representations},
  year={2020}
}

@inproceedings{engstrom2019implementation,
  title={Implementation matters in deep rl: A case study on ppo and trpo},
  author={Engstrom, L. and Ilyas, A. and Santurkar, S. and Tsipras, D. and Janoos, F. and Rudolph, L. and Madry, A.},
  booktitle={Int. Conf. learning representations},
  year={2019}
}

@inproceedings{perez2018film,
  title={Film: Visual reasoning with a general conditioning layer},
  author={Perez, E. and Strub, F. and De Vries, H. and Dumoulin, V. and Courville, A.},
  booktitle={Proc AAAI Conf. Artificial intelligence},
  volume={32},
  number={1},
  year={2018}
}

@article{rockafellar2000optimization,
  title={Optimization of conditional value-at-risk},
  author={Rockafellar, R Tyrrell and Uryasev, Stanislav and others},
  journal={Journal of risk},
  volume={2},
  pages={21--42},
  year={2000}
}

@article{dixit2023risk,
  title={Risk-averse receding horizon motion planning for obstacle avoidance using coherent risk measures},
  author={Dixit, Anushri and Ahmadi, Mohamadreza and Burdick, Joel W},
  journal={Artificial Intelligence},
  volume={325},
  pages={104018},
  year={2023},
  publisher={Elsevier}
}

@inproceedings{handa2023dextreme,
  title={Dextreme: Transfer of agile in-hand manipulation from simulation to reality},
  author={Handa, A. and Allshire, A. and Makoviychuk, V. and Petrenko, A. and Singh, R. and Liu, J. and Makoviichuk, D. and Van Wyk, K. and Zhurkevich, A. and Sundaralingam, B. et. al},
  booktitle={IEEE Int. Conf. Robotics and Automation},
  pages={5977--5984},
  year={2023}
}

@inproceedings{tobin2017domain,
  title={Domain randomization for transferring deep neural networks from simulation to the real world},
  author={Tobin, J. and Fong, R. and Ray, A. and Schneider, J. and Zaremba, W. and Abbeel, P.},
  booktitle={IEEE/RSJ Int. Conf. intelligent robots and systems},
  pages={23--30},
  year={2017}
}

@article{chi2024universal,
  title={Universal manipulation interface: In-the-wild robot teaching without in-the-wild robots},
  author={Chi, C. and Xu, Z. and Pan, C. and Cousineau, E. and Burchfiel, B. and Feng, S. and Tedrake, R. and Song, S.},
  journal={arXiv:2402.10329},
  year={2024}
}

@article{suh2022bundled,
  title={Bundled gradients through contact via randomized smoothing},
  author={Suh, Hyung Ju Terry and Pang, Tao and Tedrake, Russ},
  journal={IEEE Robotics and Automation Letters},
  volume={7},
  number={2},
  pages={4000--4007},
  year={2022},
  publisher={IEEE}
}

@article{le2024leveraging,
  title={Leveraging randomized smoothing for optimal control of nonsmooth dynamical systems},
  author={Le Lidec, Quentin and Schramm, Fabian and Montaut, Louis and Schmid, Cordelia and Laptev, Ivan and Carpentier, Justin},
  journal={Nonlinear Analysis: Hybrid Systems},
  volume={52},
  pages={101468},
  year={2024},
  publisher={Elsevier}
}

@article{wensing2023optimization,
  title={Optimization-based control for dynamic legged robots},
  author={Wensing, Patrick M and Posa, Michael and Hu, Yue and Escande, Adrien and Mansard, Nicolas and Del Prete, Andrea},
  journal={IEEE Transactions on Robotics},
  year={2023},
  publisher={IEEE}
}

@article{vlahov2024mppi,
  title={Mppi-generic: A cuda library for stochastic optimization},
  author={Vlahov, B. and Gibson, J. and Gandhi, M. and Theodorou, E.A.},
  journal={  arXiv:2409.07563},
  year={2024}
}

@article{wang2021variational,
  title={Variational inference MPC using Tsallis divergence},
  author={Wang, Z. and So, O. and Gibson, J. and Vlahov, B. and Gandhi, M.S. and Liu, G.-H. and Theodorou, E.A.},
  journal={arXiv:2104.00241},
  year={2021}
}

@inproceedings{gao2025diffusionmeetsflow,
  author = {Gao, R. and Hoogeboom, E. and Heek, J. and Bortoli, V.D. and Murphy, K.P. and Salimans, T.},
  title = {Diffusion Meets Flow Matching: Two Sides of the Same Coin},
  year = {2024},
  url  = {https://diffusionflow.github.io/}
}

@article{team2025gemini,
  title={Gemini Robotics: Bringing AI into the Physical World},
  author={Gemini Robotics Team},
  journal={arXiv:2503.20020},
  year={2025}
}

@article{zhu2024should,
  title={Should We Learn Contact-Rich Manipulation Policies from Sampling-Based Planners?},
  author={Zhu, H. and Zhao, T. and Ni, X. and Wang, J. and Fang, K. and Righetti, L. and Pang, T.},
  journal={  arXiv:2412.09743},
  year={2024}
}

@article{schulman2017proximal,
  title={Proximal policy optimization algorithms},
  author={Schulman, John and Wolski, Filip and Dhariwal, Prafulla and Radford, Alec and Klimov, Oleg},
  journal={arXiv preprint arXiv:1707.06347},
  year={2017}
}

@article{black2025real,
  title={Real-Time Execution of Action Chunking Flow Policies},
  author={Black, Kevin and Galliker, Manuel Y and Levine, Sergey},
  journal={arXiv preprint arXiv:2506.07339},
  year={2025}
}
